\documentclass{article} 
\usepackage{iclr2027_conference,times}

\usepackage{amsmath,amsfonts,bm}

\def\eqref#1{equation~\ref{#1}}

\def\1{\bm{1}}

\DeclareMathAlphabet{\mathsfit}{\encodingdefault}{\sfdefault}{m}{sl}
\SetMathAlphabet{\mathsfit}{bold}{\encodingdefault}{\sfdefault}{bx}{n}

\usepackage{hyperref}
\usepackage{url}
\usepackage{xcolor}
\usepackage{graphicx}
\usepackage{booktabs}
\usepackage{array}
\usepackage{multirow}
\usepackage{colortbl}
\usepackage{wrapfig}
\usepackage{placeins}
\usepackage{needspace}
\AddToHook{env/table/begin}{\setlength{\abovecaptionskip}{0pt}\setlength{\belowcaptionskip}{4pt}}
\AddToHook{env/table*/begin}{\setlength{\abovecaptionskip}{0pt}\setlength{\belowcaptionskip}{4pt}}
\AddToHook{env/wraptable/begin}{\setlength{\abovecaptionskip}{0pt}\setlength{\belowcaptionskip}{4pt}}

\title{RegionAgent: Two-Stage Preference\\
Optimization for Regional Image Restoration}

\iclrfinalcopy
\newcommand{\preprintauthors}{%
  \begin{minipage}[t]{\dimexpr\textwidth-2\tabcolsep\relax}
    \centering
    \setlength{\tabcolsep}{0pt}
    \begin{tabular}{*{4}{>{\centering\arraybackslash\bfseries}p{0.24\linewidth}}}
      Weiyu Zhou\textsuperscript{1} & Jiacong Tang\textsuperscript{1} & Tao Hu\textsuperscript{1} & Axi Niu\textsuperscript{1} \\[3pt]
      Jinqiu Sun\textsuperscript{1} & Peng Wu\textsuperscript{1} & Wei Dong\textsuperscript{2} & Qingsen Yan\textsuperscript{1,*}
    \end{tabular}\par
    \vspace{6pt}
    {\normalfont\small
      \textsuperscript{1}\,Northwestern Polytechnical University\\[1pt]
      \textsuperscript{2}\,Xi'an University of Architecture and Technology\\[2pt]
      \textsuperscript{*}\,Corresponding author.\par}
  \end{minipage}}
\author{\preprintauthors}
\hypersetup{pdftitle={RegionAgent: Two-Stage Preference Optimization for Regional Image Restoration},
  pdfauthor={Weiyu Zhou, Jiacong Tang, Tao Hu, Axi Niu, Jinqiu Sun, Peng Wu, Wei Dong, Qingsen Yan},hidelinks}

\begin{document}

\maketitle
\lhead{Preprint}

\begin{abstract}
Images under non-uniform coupled degradations require diverse restoration capabilities for different degraded regions. Existing methods employ globally uniform operations, potentially overlooking regional differences and causing incomplete restoration or even damage to image content. Moreover, a degraded region may require successive restoration actions, and earlier actions may alter residual degradation characteristics or introduce artifacts, potentially hindering subsequent restoration. To address these challenges, we propose RegionAgent, an adaptive restoration framework that trains a Vision-Language Model (VLM) through two-stage preference learning to generate a region-specific restoration action at each iteration. Specifically, we first train the VLM to select an action suited to the current region through step-level Direct Preference Optimization (DPO), using preferences derived from immediate restoration feedback to supervise only the differing component in each action pair. To account for the effects of current actions on subsequent restoration, we then apply trajectory-level DPO, using preferences derived from complete trajectory scores to supervise the sole differing action in each trajectory pair. Furthermore, as conventional image-pair datasets do not directly provide preferences over regional restoration decisions, we construct a regional restoration preference dataset comprising 10K action pairs and 20K trajectory pairs for two-stage VLM training. Experiments across multiple datasets demonstrate strong restoration performance under both mixed and single degradations. Ablation studies further validate the effectiveness of two-stage preference learning in improving restoration quality.

\end{abstract}

\section{Introduction}
\label{sec:introduction}

\begin{figure}[!t]
    \centering
    \includegraphics[pagebox=cropbox,width=\linewidth]{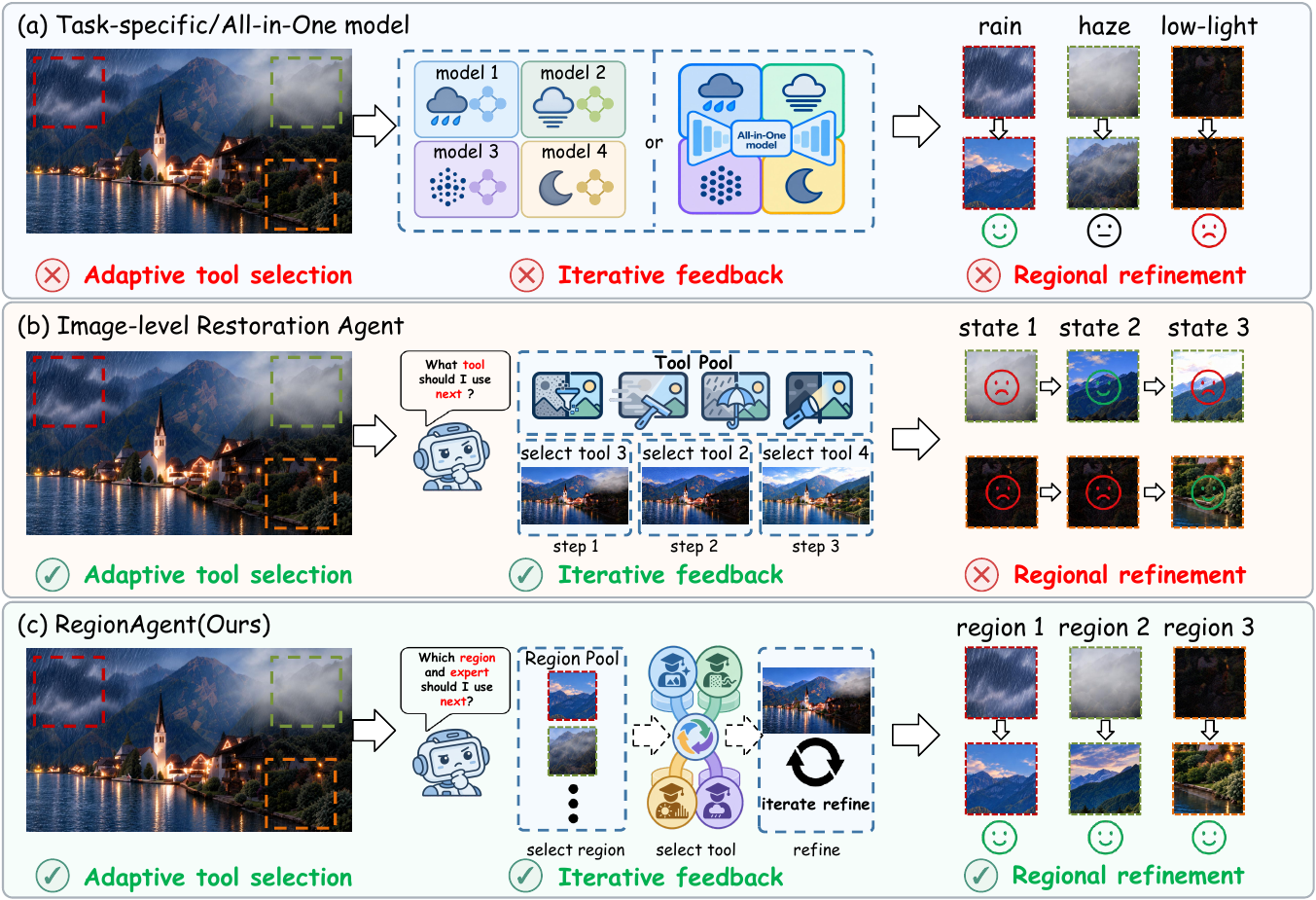}
    \caption{Conceptual comparison of restoration paradigms. (a) Task-specific and all-in-one models restore the whole image. (b) Image-level agents select tools using iterative feedback. (c) RegionAgent selects regions and expert configurations for adaptive local refinement.}
    \label{fig:restoration_paradigms}
\end{figure}

Image restoration aims to recover clean images from observations affected by degradations such as blur, haze, rain, or low light. Deep learning-based methods have advanced this task through specialized models for individual degradation types \citep{ren2021deamnet,zamir2021mprnet,yan2025hvi} and all-in-one networks for multiple restoration tasks \citep{li2022airnet,potlapalli2023promptir} (Figure~\ref{fig:restoration_paradigms}(a)). However, real-world images can contain coupled degradations that vary in type and severity across regions. Applying uniform restoration operations to the whole image may overlook regional differences and damage unaffected content. Within an affected region, removing one degradation can leave others unresolved, requiring successive restoration operations. The effectiveness of later operations then depends on how earlier ones change the image \citep{chen2024restoreagent,zhu2025agenticir,zhu2026opera}.

To make restoration more adaptive, instruction-guided methods allow users to specify restoration goals \citep{conde2024instructir} and target regions \citep{liu2025instructrestore}. Restoration agents automate tool selection and revise image-level plans using restoration feedback \citep{chen2024restoreagent,zhu2025agenticir,lin2025jarvisir} (Figure~\ref{fig:restoration_paradigms}(b)). As an alternative to sequential tool invocation, Beyond Sequential Tools learns image-level expert weights through preference optimization for single-pass restoration \citep{xiong2026beyond}. However, these approaches focus on user-specified regions or image-level decisions, rather than learning to select regions and configure their restoration autonomously.

\begin{figure*}[t!]
    \centering
    \includegraphics[pagebox=cropbox,width=\textwidth]{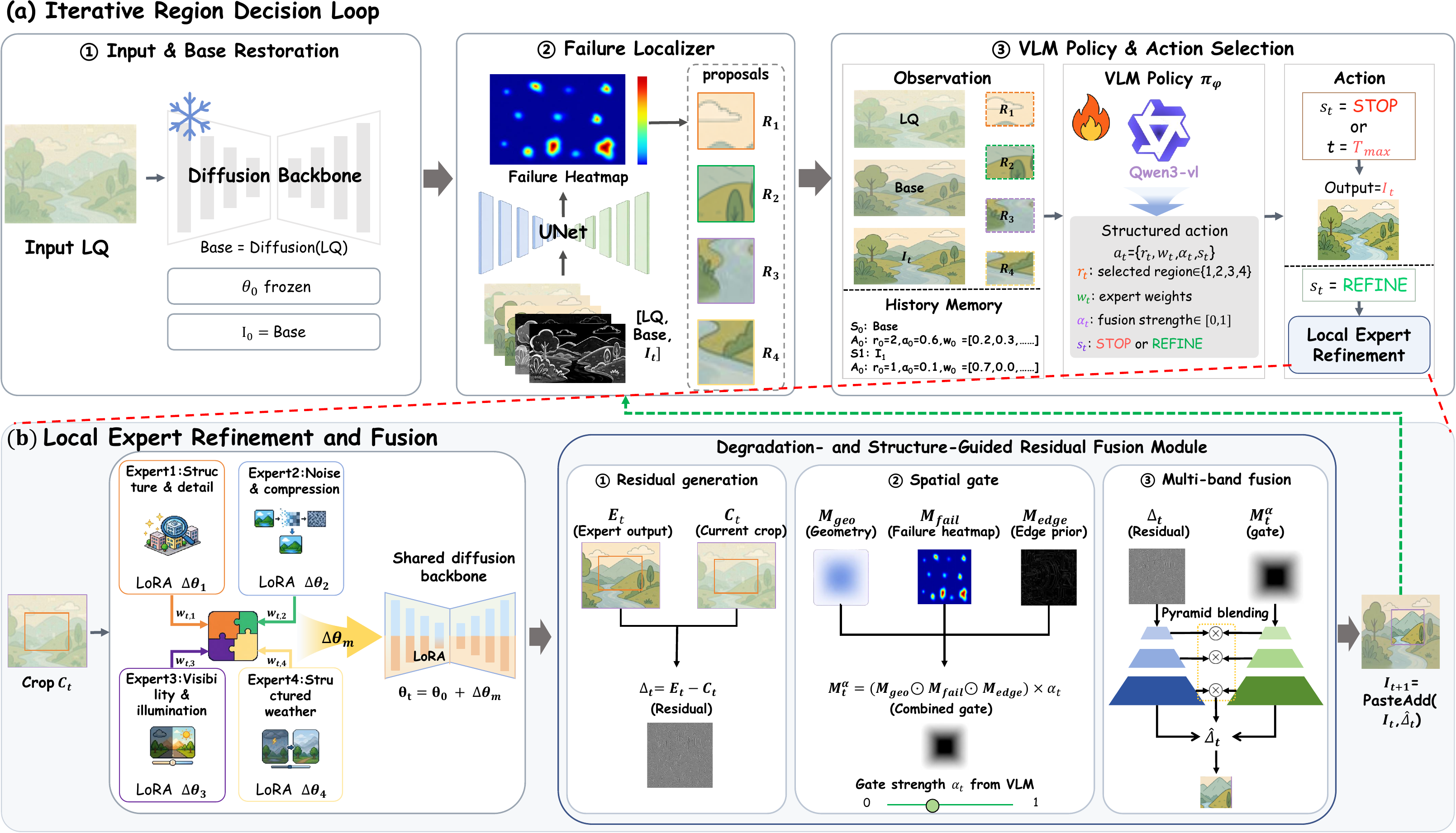}
    \caption{Overview of RegionAgent. (a) The Localizer proposes candidate regions, and the VLM selects restoration actions from the current image state. (b) The composed restorer processes the selected crop, and DSRF blends its residual correction into the image for subsequent refinement.}
    \label{fig:regionagent_inference}
\end{figure*}

We therefore introduce \emph{RegionAgent}, an adaptive restoration framework in which a Vision-Language Model (VLM) selects regions and configures their restoration from the current image state (Figure~\ref{fig:restoration_paradigms}(c)). At each iteration, a localization network predicts a heatmap of residual degradations and uses it to propose candidate regions (Figure~\ref{fig:regionagent_inference}). The VLM then selects one of these regions and specifies its restoration configuration based on the current image.

Each refinement action specifies a target region, weights for combining restoration experts, and the strength of the local correction. We train the VLM through two stages of Direct Preference Optimization (DPO) \citep{rafailov2023dpo}: step-level DPO followed by trajectory-level DPO. Step-level DPO learns which configuration choices suit the current region by comparing their immediate restoration effects, applying the preference loss only to the component that differs between paired actions. Trajectory-level DPO then trains the VLM to account for the effects of current choices on subsequent restoration. It uses complete-trajectory scores to supervise the sole differing action between paired trajectories. Since conventional image-pair datasets lack preferences over regional restoration decisions, we construct a dataset of action and trajectory preference pairs for two-stage training.

Executing these actions produces restored crops whose direct writeback can introduce boundary artifacts and structural discontinuities. Our Degradation- and Structure-Guided Residual Fusion Module (DSRF) integrates the difference between each restored crop and its current counterpart as a residual correction. Spatial masks regulate this correction according to restoration needs and image structure, while Laplacian pyramid blending promotes smooth transitions across scales. The VLM then generates its next action from the fused image.

Experiments across multiple datasets demonstrate strong restoration performance under both mixed and single degradations. Our contributions can be summarized as follows:
\begin{itemize}
    \item We introduce RegionAgent, an adaptive restoration framework whose VLM generates regional actions from the current image state to address non-uniform coupled degradations.
    \item We develop two-stage DPO: step-level DPO adapts restoration configurations to current regional needs, while trajectory-level DPO uses complete-trajectory feedback to guide current restoration actions.
    \item We construct a regional restoration preference dataset comprising 10K action pairs and 20K trajectory pairs. It includes controlled comparisons that vary a single configuration component or a single trajectory action, providing targeted supervision for regional restoration decisions beyond conventional degraded--clean image pairs.
\end{itemize}

\section{Related Work}

\paragraph{Adaptive Image Restoration.}
All-in-one models condition shared restoration networks on degradation representations and learned prompts \citep{li2022airnet,potlapalli2023promptir}, while vision-language approaches incorporate visual-textual features for degradation diagnosis and restoration guidance \citep{luo2024daclip,ai2024mperceiver,jiang2024autodir}. Instruction-guided methods extend this control to user-defined goals \citep{conde2024instructir} and region-specific restoration \citep{liu2025instructrestore}. Restoration agents automate tool selection and sequencing \citep{yu2018crafting}, using multimodal diagnosis and quality feedback to revise their plans \citep{chen2024restoreagent,zhu2025agenticir,lin2025jarvisir}. Recent systems integrate restoration with assessment \citep{chen2026rar} or couple planners with LoRA-based tools \citep{li2026hybridagent,liu2026fapeir}, while OPERA jointly optimizes planning and tool execution using restoration rewards \citep{zhu2026opera}. As an alternative to sequential execution, Beyond Sequential Tools learns image-level expert weights for single-pass restoration through a DPO-trained allocation branch \citep{xiong2026beyond}. RegionAgent instead selects regions and configures their restoration autonomously from the updated image at each iteration.

\Needspace{4\baselineskip}
\paragraph{Preference Optimization.}
Preference optimization applies to restoration outputs and decision policies. DSPO aligns super-resolution generators with semantic-instance preferences \citep{cai2025dspo}, whereas TTPO guides diffusion sampling with test-time preferences without model training \citep{li2026ttpo}. Within policy outputs, fine-grained objectives weight tokens or target localized differences \citep{christopoulou2025sparsepo,hasan2025teaching}, and DiTTo separates preference supervision across degradation reasoning, restoration order, and tool calls \citep{choi2026ditto}. For sequential decisions, Hierarchical Preference Learning combines trajectory-, step-, and action-group objectives \citep{gao2026hpl}, while Critical Step Optimization uses successful policy continuations to verify alternative actions for action-level DPO \citep{li2026cso}. Our refinement pairs vary one configuration component from a shared image state; trajectory pairs vary one action while fixing the region and other actions. The losses target the differing component and full action, respectively.

\section{Method}
\label{sec:method}

\subsection{Problem Formulation}
\label{sec:problem_formulation}

Given a low-quality image $I_{\mathrm{LQ}}$ of height $H$ and width $W$, a pretrained shared restorer $E_{\theta_0}$ with frozen parameters $\theta_0$ produces the fixed reference $I_{\mathrm{Base}}=E_{\theta_0}(I_{\mathrm{LQ}})$ and initializes the current image as $I_0=I_{\mathrm{Base}}$. At iteration $t$, the VLM selects an action $a_t\in\{\texttt{STOP}\}\cup\{\texttt{REFINE}(r_t,\mathbf{w}_t,\alpha_t)\}$ from the current restoration state. A refinement action specifies a region $r_t$ from the candidate region set $\mathcal{R}_t$, expert combination weights $\mathbf{w}_t$, and fusion strength $\alpha_t$, with the latter two taking discrete values. The process returns the current image upon \texttt{STOP} or at the iteration limit $T_{\max}$.

We first describe regional decisions and execution (Secs.~\ref{sec:closed_loop_decision}--\ref{sec:regional_refinement}), then preference construction and two-stage DPO for learning these decisions (Secs.~\ref{sec:supervision_construction}--\ref{sec:preference_optimization}).

\Needspace{5\baselineskip}
\subsection{Iterative Regional Decision-Making}
\label{sec:closed_loop_decision}

We train a lightweight U-Net Localizer $f_{\omega}$, with parameters $\omega$, to localize residual degradations. To cover different restoration stages, we derive training targets from discrepancies between degraded inputs or intermediate restoration results and their corresponding clean images.

\Needspace{7\baselineskip}
After training, we freeze $\omega$ and predict the failure heatmap $F_t=f_{\omega}(X_t)\in[0,1]^{H\times W}$ at each iteration from the input tensor $X_t=\operatorname{cat}[I_{\mathrm{LQ}},I_{\mathrm{Base}},I_t,|I_{\mathrm{LQ}}-I_t|,|I_{\mathrm{Base}}-I_t|]$, where $I_t$ is the current image and $\operatorname{cat}$ denotes channel-wise concatenation. Higher heatmap values indicate a greater need for further restoration. We rank square windows of different sizes by their mean heatmap values and apply non-maximum suppression to select four new proposals. The candidate region set $\mathcal{R}_t$ contains these proposals and, for $t>0$, the previously selected region $r_{t-1}$, with duplicate boxes included once.

Given the observation $o_t=(I_{\mathrm{LQ}},I_{\mathrm{Base}},I_t,\mathcal{R}_t,\mathcal{H}_t)$, the VLM policy $\pi_{\varphi}$ with parameters $\varphi$ selects $a_t\sim\pi_{\varphi}(\cdot\mid o_t)$, where $\mathcal{H}_t$ is the structured-action history. After refinement produces $I_{t+1}$, we update the heatmap and proposals while retaining $r_t$, allowing the VLM to continue refining it or select another region (Figure~\ref{fig:regionagent_inference}(a)).

\subsection{Regional Refinement}
\label{sec:regional_refinement}

To process candidate regions of different sizes at a common input resolution, we construct a crop $C_t$ of size $s_p\times s_p$. The selected region $r_t$ occupies a central core of size $s_c\times s_c$, while the surrounding halo provides structural and textural context and a transition area for fusion. An affine transform $B_t$ maps $r_t$ to this core. Using the same transform, the sampling operator $\operatorname{Sample}$ extracts the current-image crop $C_t=\operatorname{Sample}(I_t;B_t,s_p)$ and the spatially aligned failure map $\widetilde F_t=\operatorname{Sample}(F_t;B_t,s_p)$.

Each expert is a LoRA update $\Delta\theta_k$ to the shared backbone \citep{hu2022lora}. The executor combines them as $\Delta\theta_m=\sum_k w_{t,k}\Delta\theta_k$ \citep{xiong2026beyond}, where $k$ indexes experts, $w_{t,k}$ is the corresponding VLM-selected weight in $\mathbf w_t$, and $m$ denotes the mixture. The composed parameters $\theta_t=\theta_0+\Delta\theta_m$ produce the restored crop $E_t=E_{\theta_t}(C_t)$ in one forward pass. The residual $\Delta_t=E_t-C_t$ lets fusion control where and how strongly this correction changes the current image.

Directly adding the crop residual can unnecessarily alter well-restored pixels and create abrupt transitions to surrounding content. We therefore design the \emph{Degradation- and Structure-Guided Residual Fusion Module} (DSRF), which combines three masks to regulate corrections according to spatial position, predicted restoration need, and local structural changes. The geometric mask $M_{\mathrm{geo}}$ equals one in the core and smoothly decays to zero through the halo, providing a gradual transition to surrounding content. The failure mask $M_{\mathrm{fail}}$ weights corrections by the restoration need predicted in $\widetilde F_t$, using $C_t$-guided filtering to align mask boundaries with image structure \citep{he2013guided}. The edge mask $M_{\mathrm{edge}}$ further reduces correction weights at halo pixels with large gradient-magnitude differences between $E_t$ and $C_t$, while remaining one in the core to permit structural correction. Together, these masks define the spatial gate $M_t^{\alpha}=\alpha_t(M_{\mathrm{geo}}\odot M_{\mathrm{fail}}\odot M_{\mathrm{edge}})$, where $\alpha_t$ controls fusion strength, each mask lies in $[0,1]$, and $\odot$ denotes element-wise multiplication.

To blend appearance and detail across scales, DSRF uses this spatial gate to weight the Laplacian pyramid of the crop residual and reconstructs the fused residual $\widehat{\Delta}_t=\operatorname{Recon}\{G_{\ell}(M_t^{\alpha})\odot L_{\ell}(\Delta_t)\}_{\ell=0}^{J-1}$. Here, $L_{\ell}$ and $G_{\ell}$ extract Laplacian and Gaussian pyramid levels indexed by $\ell$, $J$ is the number of levels, and $\operatorname{Recon}$ reconstructs the pyramid \citep{burt1983multiresolution}. Residual writeback gives $I_{t+1}=\operatorname{PasteAdd}(I_t,\widehat{\Delta}_t)=I_t+\mathcal{W}_{B_t}^{-1}(\widehat{\Delta}_t)$, where $\mathcal{W}_{B_t}^{-1}$ maps the crop residual back to the image with zero support outside the sampled context.

\begin{figure}[!htbp]
    \centering
    \includegraphics[pagebox=cropbox,width=\linewidth]{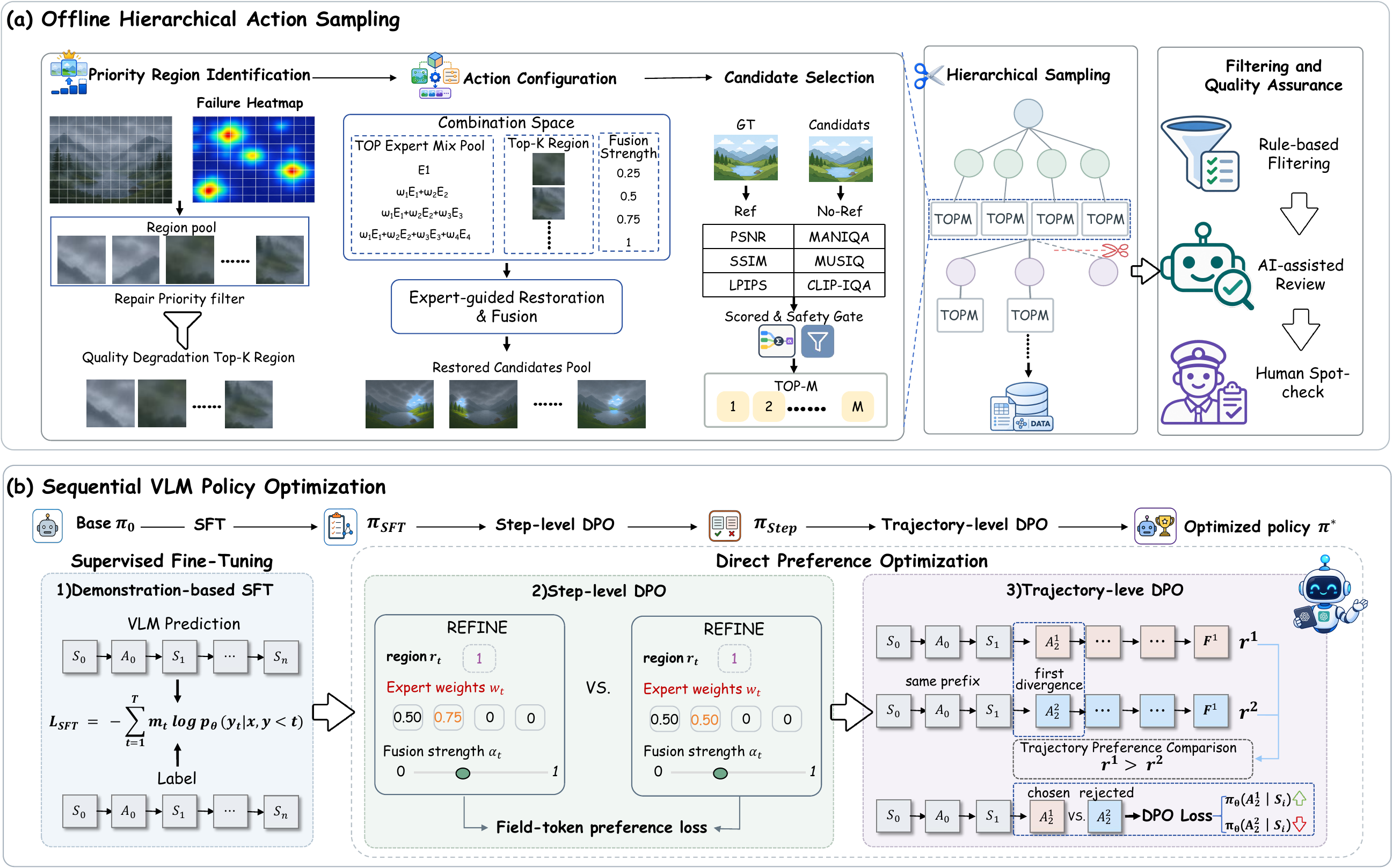}
    \caption{Supervision construction and policy optimization. (a) Action sampling, scoring, trajectory expansion, and filtering produce demonstrations and preference pairs. (b) Supervised fine-tuning and two-stage DPO train regional decisions.}
    \label{fig:supervision_pipeline}
\end{figure}

\subsection{Supervision Construction}
\label{sec:supervision_construction}

Conventional image-pair datasets provide degraded images and clean targets, but do not directly provide preferences over regional restoration decisions. We therefore construct a regional restoration preference dataset containing action and trajectory pairs to support two-stage DPO.

To control search cost, we first restore each newly proposed region with one expert at a time, with every trial starting from the same current image. We fuse each restored crop at the allowed fusion strengths and score the resulting images using the one-step utility $U_{\mathrm{step}}$, which measures restoration gains relative to the current image after accounting for penalties. We rank regions by their highest utility across these trials and expand expert configurations in the top two regions.

The one-step utility $U_{\mathrm{step}}$ uses three full-reference metrics (PSNR, SSIM, and LPIPS) and three no-reference metrics (MANIQA, MUSIQ, and CLIP-IQA). We normalize and clip each metric gain separately. For full-reference metrics, we equally combine full-image and selected-region gains, whereas no-reference metrics use full-image gains. We average the resulting gains across the six metrics and subtract penalties for halo degradation, boundary artifacts, action cost, repeated states, and degradation below Base.

Beam search then builds complete restoration sequences with region coordinates held fixed, updating successor heatmaps and admitting successors only when their one-step utilities exceed a minimum threshold. This fixed-region constraint applies to offline trajectory construction, while inference selects regions dynamically (Sec.~\ref{sec:closed_loop_decision}). For a trajectory $\tau_t$ ending at $I_t$ after $t$ refinement steps, we define its normalized quality gain $Q(I_t)$ and penalized trajectory score $r(\tau_t)$ as
\begin{equation*}
\begin{aligned}
Q(I_t)&=\frac{1}{|\mathcal{M}|}\sum_{m\in\mathcal{M}}
\operatorname{clip}\!\left(
\frac{d_m[m(I_t)-m(I_{\mathrm{Base}})]}{s_m+\epsilon},-1,1\right),\\
r(\tau_t)&=Q(I_t)-\lambda_{\mathrm{out}}\sum_{i<t}P_i^{\mathrm{out}}
-\lambda_{\mathrm{bnd}}\sum_{i<t}P_i^{\mathrm{bnd}}
-\lambda_{\mathrm{len}}t-\lambda_{\mathrm{rep}}N_{\mathrm{rep}}-\lambda_{\mathrm{base}}P_{\mathrm{base}}.
\end{aligned}
\end{equation*}
Here, $\mathcal M$ contains the six metrics above, and $m(I)$ is the full-image score for metric $m$, using the corresponding clean image for full-reference metrics. The sign $d_m\in\{-1,+1\}$ aligns improvement, $s_m$ is the metric's normalization scale, and $\epsilon>0$ ensures numerical stability.

At step $i$, $P_i^{\mathrm{out}}$ averages positive increases in absolute error to the clean image over the halo, the band inside the writeback support but outside the core. Similarly, $P_i^{\mathrm{bnd}}$ penalizes increased gradient error to the clean image near the core boundary. The term $N_{\mathrm{rep}}$ counts repeated states along the trajectory, and $P_{\mathrm{base}}$ penalizes degradation of the final image relative to Base. The $\lambda$ terms weight these penalties and refinement count $t$. This score compares trajectories ending with \texttt{STOP} or reaching $T_{\max}$.

We form step-level pairs from same-state actions. To construct a trajectory pair, we replace one refinement action in a completed search trajectory while keeping its target region fixed, then execute the unchanged subsequent actions from the resulting image state. The original and modified trajectories therefore differ in exactly one action, while subsequent image states may differ. Filtering enforces pairing constraints and executability, removes duplicates, and balances degradation and decision coverage. Pairs meeting score-gap thresholds are selected automatically; low-margin or metric-conflicting cases undergo VLM-assisted screening and human verification. The best trajectories form the demonstration set $\mathcal{D}_{\mathrm{SFT}}$ for supervised fine-tuning (SFT), and retained action and trajectory pairs form $\mathcal{D}_{\mathrm{step}}$ and $\mathcal{D}_{\mathrm{traj}}$.

\subsection{Preference Optimization}
\label{sec:preference_optimization}

SFT first teaches the VLM to generate correctly formatted, executable actions from demonstrations. Step-level and trajectory-level DPO then train regional configuration choices and decisions that account for subsequent restoration, respectively.

Both DPO stages use a common objective with stage-specific labels and token masks. For action $a$ under observation $o$, let $y_j$ denote its $j$-th serialized token, $y_{<j}$ its full token prefix, and $\mathcal T(a)$ the supervised positions. The selected-token log-probability ratio $z_\pi^{\mathcal T}$ and DPO loss for preferred/rejected actions $(a^+,a^-)$ are
\begin{align*}
z_{\pi}^{\mathcal{T}}(a\mid o)
&=\sum_{j\in\mathcal{T}(a)}
\log\frac{\pi(y_j\mid o,y_{<j})}
{\pi_{\mathrm{ref}}(y_j\mid o,y_{<j})},\\
\mathcal{L}_{\mathrm{DPO}}(\mathcal{T})
&=-\mathbb{E}\log\sigma\!\left(
\beta[z_{\pi}^{\mathcal{T}}(a^+\mid o)
-z_{\pi}^{\mathcal{T}}(a^-\mid o)]\right),
\end{align*}
where $\pi=\pi_\varphi$ is the trainable VLM policy, $\pi_{\mathrm{ref}}$ is its frozen reference, $\beta$ controls preference strength, and $\sigma$ is the logistic function \citep{rafailov2023dpo}. We sum selected-token log-probability ratios without length normalization. The expectation averages retained pairs from the current stage's dataset; trajectory pairs supply the shared observation and alternative actions defined below.

Step-level DPO learns configurations suited to current regional needs from paired \texttt{REFINE} actions that share an observation and differ in one field: region, expert combination weights, or fusion strength. Their one-step utilities determine the preference, and $\mathcal{T}$ selects that field's value tokens, including the full expert-weight vector for weight comparisons. For \texttt{STOP}--\texttt{REFINE} pairs, complete-trajectory scores compare immediate termination with continued restoration, while $\mathcal{T}$ selects only the \texttt{action} value tokens. The frozen SFT checkpoint supplies the reference policy.

Trajectory-level DPO evaluates a current decision through its effects on subsequent restoration of the same region. For $(\tau^+,\tau^-)\in\mathcal{D}_{\mathrm{traj}}$ with $r(\tau^+)>r(\tau^-)$, let $t^{\dagger}$ be their sole differing iteration, so $a_t^+=a_t^-$ for $t\ne t^{\dagger}$. The shared prefix supplies the same observation $o_{t^{\dagger}}$, and $\mathcal{T}$ selects every token of the alternative actions at $t^{\dagger}$. With the region and remaining action configurations held fixed, complete-trajectory scores provide supervision for the differing action's effect on subsequent refinement. The trainable and frozen reference policies both start from the step-level checkpoint. Retained pairs receive equal weight.

\section{Experiments}
\label{sec:experiments}

\begin{table}[t]
  \centering
  \caption{Quantitative comparison on MiO100. Base is the shared restorer's output before regional refinement. Best and second-best results per group are shown in \textbf{bold} and \underline{underlined}.}
  \label{tab:mio100_main}
  \footnotesize
  \setlength{\tabcolsep}{2.8pt}
  \renewcommand{\arraystretch}{0.86}
  \resizebox{\textwidth}{!}{%
  \begin{tabular}{@{}clc*{6}{c}@{}}
    \toprule
    \textbf{Degradations} & \textbf{Method} & \textbf{Venue} & \textbf{PSNR}$\uparrow$ & \textbf{SSIM}$\uparrow$ & \textbf{LPIPS}$\downarrow$ & \textbf{MANIQA}$\uparrow$ & \textbf{CLIP-IQA}$\uparrow$ & \textbf{MUSIQ}$\uparrow$ \\
    \midrule
    \multirow[c]{14}{*}{Group A}
      & AirNet~\citeyearpar{li2022airnet}               & CVPR    & 19.13 & 0.6019 & 0.4283 & 0.2581 & 0.3930 & 42.46 \\
      & PromptIR~\citeyearpar{potlapalli2023promptir}   & NeurIPS & 20.06 & 0.6088 & 0.4127 & 0.2633 & 0.4013 & 42.62 \\
      & MiOIR~\citeyearpar{kong2024mioir}               & arXiv   & 20.84 & 0.6558 & 0.3715 & 0.2451 & 0.3933 & 47.82 \\
      & DA-CLIP~\citeyearpar{luo2024daclip}             & ICLR    & 19.58 & 0.6032 & 0.4266 & 0.2418 & 0.4139 & 42.51 \\
      & InstructIR~\citeyearpar{conde2024instructir}    & ECCV    & 18.03 & 0.5751 & 0.4429 & 0.2660 & 0.3528 & 45.77 \\
      & AdaIR~\citeyearpar{cui2025adair}                & ICLR    & 19.06 & 0.6061 & 0.4170 & 0.2718 & 0.4117 & 44.66 \\
      & DA-RCOT~\citeyearpar{tang2025darcot}            & TPAMI   & 18.82 & 0.5960 & 0.4303 & 0.2641 & 0.4058 & 43.42 \\
      & AutoDIR~\citeyearpar{jiang2024autodir}          & ECCV    & 19.64 & 0.6286 & 0.3967 & 0.2500 & 0.3767 & 47.01 \\
      & AgenticIR~\citeyearpar{zhu2025agenticir}        & ICLR    & 21.04 & 0.6818 & 0.3148 & 0.3071 & 0.4474 & 56.88 \\
      & MAIR~\citeyearpar{jiang2026mair}                & IJCV    & 21.02 & 0.6715 & 0.2963 & 0.3330 & 0.4751 & 59.19 \\
      & 4KAgent~\citeyearpar{zuo20254kagent}            & NeurIPS & \underline{21.48} & 0.6720 & 0.3019 & 0.3748 & 0.5544 & \underline{63.19} \\
      & RAR~\citeyearpar{chen2026rar}                   & CVPR    & 20.46 & \underline{0.7144} & \textbf{0.1299} & \underline{0.4659} & \textbf{0.6566} & 57.19 \\
      & Base (Ours) & Ours & 21.34 & 0.6837 & 0.2916 & 0.4318 & 0.5417 & 62.38 \\
    \rowcolor{gray!12}
      & \textbf{RegionAgent (Ours)}                & Ours         & \textbf{22.11} & \textbf{0.7150} & \underline{0.2377} & \textbf{0.5834} & \underline{0.5652} & \textbf{67.41} \\
    \midrule
    \multirow[c]{14}{*}{Group B}
      & AirNet~\citeyearpar{li2022airnet}               & CVPR    & 19.31 & 0.6567 & 0.3670 & 0.2882 & 0.4274 & 47.88 \\
      & PromptIR~\citeyearpar{potlapalli2023promptir}   & NeurIPS & 20.47 & 0.6704 & 0.3370 & 0.2893 & 0.4289 & 48.10 \\
      & MiOIR~\citeyearpar{kong2024mioir}               & arXiv   & 20.56 & 0.6905 & 0.3243 & 0.2638 & 0.4330 & 51.87 \\
      & DA-CLIP~\citeyearpar{luo2024daclip}             & ICLR    & 18.56 & 0.5946 & 0.4405 & 0.2435 & 0.4154 & 43.70 \\
      & InstructIR~\citeyearpar{conde2024instructir}    & ECCV    & 18.34 & 0.6235 & 0.4072 & 0.3022 & 0.3790 & 50.94 \\
      & AdaIR~\citeyearpar{cui2025adair}                & ICLR    & 19.24 & 0.6637 & 0.3467 & 0.3001 & 0.4380 & 49.16 \\
      & DA-RCOT~\citeyearpar{tang2025darcot}            & TPAMI   & 19.08 & 0.6372 & 0.3910 & 0.2845 & 0.4321 & 46.37 \\
      & AutoDIR~\citeyearpar{jiang2024autodir}          & ECCV    & 19.90 & 0.6643 & 0.3542 & 0.2534 & 0.3986 & 49.64 \\
      & AgenticIR~\citeyearpar{zhu2025agenticir}        & ICLR    & 20.55 & 0.7009 & 0.3072 & 0.3204 & 0.4648 & 57.57 \\
      & MAIR~\citeyearpar{jiang2026mair}                & IJCV    & 20.92 & 0.7004 & 0.2788 & 0.3544 & 0.5084 & 60.98 \\
      & 4KAgent~\citeyearpar{zuo20254kagent}            & NeurIPS & 20.95 & 0.6727 & 0.3017 & 0.3734 & 0.5505 & \underline{62.69} \\
      & RAR~\citeyearpar{chen2026rar}                   & CVPR    & \underline{21.04} & \underline{0.7326} & \textbf{0.1269} & \underline{0.4582} & \textbf{0.6483} & 56.91 \\
      & Base (Ours) & Ours & 20.87 & 0.7083 & 0.2839 & 0.4265 & 0.5362 & 61.74 \\
    \rowcolor{gray!12}
      & \textbf{RegionAgent (Ours)}                & Ours         & \textbf{21.61} & \textbf{0.7496} & \underline{0.2324} & \textbf{0.5790} & \underline{0.5609} & \textbf{67.26} \\
    \midrule
    \multirow[c]{15}{*}{Group C}
      & AirNet~\citeyearpar{li2022airnet}               & CVPR    & 17.95 & 0.5145 & 0.5782 & 0.1854 & 0.3113 & 30.12 \\
      & PromptIR~\citeyearpar{potlapalli2023promptir}   & NeurIPS & 18.51 & 0.5166 & 0.5756 & 0.1906 & 0.3104 & 29.71 \\
      & MiOIR~\citeyearpar{kong2024mioir}               & arXiv   & 15.63 & 0.4896 & 0.5376 & 0.1717 & 0.2891 & 37.95 \\
      & DA-CLIP~\citeyearpar{luo2024daclip}             & ICLR    & 18.53 & 0.5320 & 0.5335 & 0.1916 & 0.3476 & 33.87 \\
      & InstructIR~\citeyearpar{conde2024instructir}    & ECCV    & 17.09 & 0.5135 & 0.5582 & 0.1732 & 0.2537 & 33.69 \\
      & AdaIR~\citeyearpar{cui2025adair}                & ICLR    & 17.88 & 0.5209 & 0.5802 & 0.2036 & 0.3276 & 30.83 \\
      & DA-RCOT~\citeyearpar{tang2025darcot}            & TPAMI   & 17.45 & 0.4993 & 0.6097 & 0.1937 & 0.3172 & 29.50 \\
      & AutoDIR~\citeyearpar{jiang2024autodir}          & ECCV    & 18.61 & 0.5443 & 0.5019 & 0.2045 & 0.2939 & 37.86 \\
      & AgenticIR~\citeyearpar{zhu2025agenticir}        & ICLR    & 18.82 & 0.5474 & 0.4493 & 0.2698 & 0.3948 & 48.68 \\
      & MAIR~\citeyearpar{jiang2026mair}                & IJCV    & 19.42 & 0.5544 & 0.4142 & 0.2798 & 0.4239 & 51.36 \\
      & 4KAgent~\citeyearpar{zuo20254kagent}            & NeurIPS & 19.77 & 0.5629 & 0.4271 & 0.3545 & \underline{0.5233} & 55.56 \\
      & RAR~\citeyearpar{chen2026rar}                   & CVPR    & 19.33 & \underline{0.6579} & \textbf{0.1489} & \underline{0.4653} & \textbf{0.6554} & 56.56 \\
      & Xiong et al.~\citeyearpar{xiong2026beyond}     & CVPR    & \underline{19.85} & 0.5765 & 0.3392 & 0.3298 & 0.4742 & \underline{58.78} \\
      & Base (Ours) & Ours & 19.68 & 0.5874 & 0.3867 & 0.3729 & 0.4938 & 56.23 \\
    \rowcolor{gray!12}
      & \textbf{RegionAgent (Ours)}                & Ours         & \textbf{20.90} & \textbf{0.6591} & \underline{0.2976} & \textbf{0.5385} & 0.5222 & \textbf{64.29} \\
    \bottomrule
  \end{tabular}%
  }
\end{table}

\subsection{Experimental Settings}
\label{sec:experimental_settings}

\textbf{Evaluation benchmarks.}
We evaluate mixed degradations on AgenticIR's MiO100-derived benchmark~\citep{zhu2025agenticir,kong2024mioir}, which applies degradation operators sequentially to each source image, and single degradations on RAR's eight-dataset suite~\citep{chen2026rar}. Tables~\ref{tab:mio100_main} and~\ref{tab:single_degradation} compare RegionAgent with all-in-one models, diffusion-based restorers, and restoration agents.

\textbf{Evaluation metrics.}
PSNR, SSIM, and LPIPS measure fidelity to the clean image, while CLIP-IQA, MUSIQ, and MANIQA provide no-reference quality estimates~\citep{wang2004ssim,zhang2018perceptual,wang2023clipiqa,ke2021musiq,yang2022maniqa}. Higher values are better except for LPIPS.

\textbf{Implementation details.}
We train the VLM Qwen3-VL-4B-Instruct~\citep{bai2025qwen3vl} with SFT followed by step-level and trajectory-level DPO. The framework combines an OSEDiff-based shared restorer~\citep{wu2024osediff}, four LoRA experts, and a frozen U-Net Localizer~\citep{ronneberger2015unet}. Inference permits at most $T_{\max}=8$ \texttt{REFINE} actions, stopping earlier on \texttt{STOP}. Experiments run on NVIDIA A100 GPUs.

\subsection{Main Results}
\label{sec:main_results}

\textbf{Quantitative comparison.}
RegionAgent achieves the highest PSNR, SSIM, MANIQA, and MUSIQ across all three mixed-degradation groups (Table~\ref{tab:mio100_main}). Relative to the shared restorer's Base output, regional refinement improves all six metrics and raises PSNR by 0.77, 0.74, and 1.22~dB on Groups A, B, and C, respectively. On Group C with three coupled degradations~\citep{zhu2025agenticir}, it reaches 20.90~dB, exceeding the strongest PSNR baseline from \citet{xiong2026beyond} by 1.05~dB.

\FloatBarrier

\Needspace{18\baselineskip}
\begin{wraptable}{r}{0.55\textwidth}
  \vspace{-4pt}
  \centering
  \caption{Average performance on single-degradation benchmarks. Best and second-best results are shown in \textbf{bold} and \underline{underlined}, respectively. LD denotes latent diffusion.}
  \label{tab:single_degradation}
  \scriptsize
  \setlength{\tabcolsep}{1.0pt}
  \renewcommand{\arraystretch}{0.92}
  \begin{tabular*}{\linewidth}{@{\extracolsep{\fill}}l*{6}{c}@{}}
    \toprule
    \textbf{Method} & \textbf{PSNR}$\uparrow$ & \textbf{SSIM}$\uparrow$ & \textbf{LPIPS}$\downarrow$ & \textbf{CLIP-IQA}$\uparrow$ & \textbf{MUSIQ}$\uparrow$ & \textbf{MANIQA}$\uparrow$ \\
    \midrule
    AirNet             & 25.65 & 0.8230 & 0.1820 & 0.4516 & 48.69 & 0.3163 \\
    PromptIR           & 27.04 & 0.8445 & 0.1515 & 0.4196 & 48.77 & 0.3169 \\
    AdaIR              & 26.48 & 0.7825 & 0.2604 & 0.4368 & 53.51 & 0.3456 \\
    DA-RCOT            & 26.08 & 0.7772 & 0.2579 & 0.4409 & 54.34 & 0.3461 \\
    LD                 & 22.17 & 0.6922 & 0.1935 & 0.4253 & 54.94 & 0.3438 \\
    AutoDIR            & \underline{27.81} & \underline{0.8703} & 0.1283 & 0.4119 & 54.37 & 0.3053 \\
    AgenticIR          & 24.41 & 0.7794 & 0.1392 & 0.3763 & 55.95 & 0.2978 \\
    RAR                & 25.88 & 0.8378 & \textbf{0.0699} & \textbf{0.5566} & \underline{56.00} & \underline{0.4125} \\
    \rowcolor{gray!12}
    \textbf{RegionAgent}       & \textbf{28.71} & \textbf{0.8861} & \underline{0.1269} & \underline{0.5313} & \textbf{58.35} & \textbf{0.6830} \\
    \bottomrule
  \end{tabular*}
  \vspace{-4pt}
\end{wraptable}

On the single-degradation suite, RegionAgent achieves an average PSNR of 28.71~dB, exceeding the strongest PSNR baseline, AutoDIR, by 0.90~dB, together with the highest average SSIM, MANIQA, and MUSIQ (Table~\ref{tab:single_degradation}).

\textbf{Inference cost.}
RegionAgent takes 18.60 seconds per image on average and uses 9.28\,GB of peak GPU memory.

\begin{figure}[!htbp]
  \centering
  \includegraphics[pagebox=cropbox,width=\linewidth]{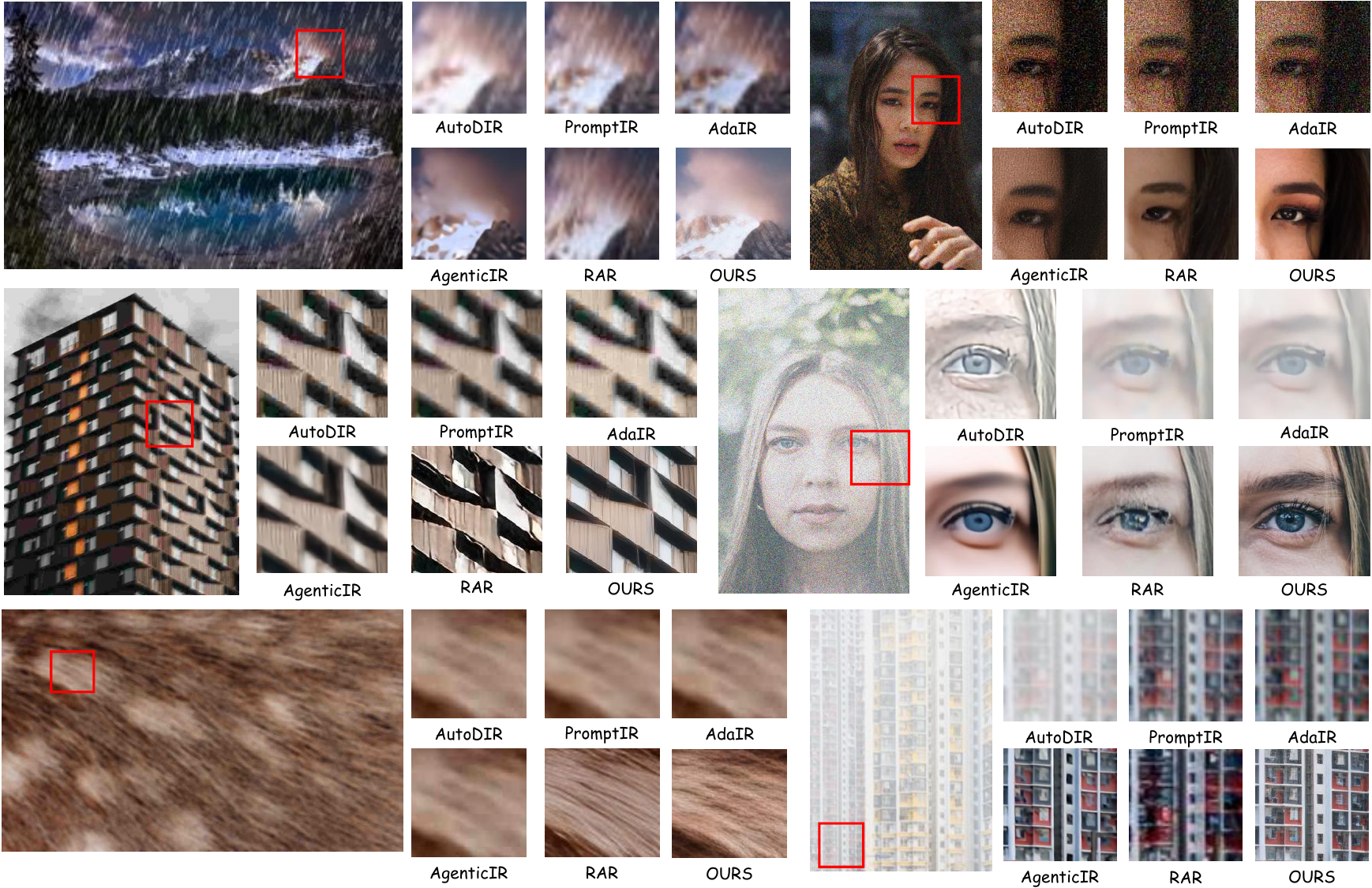}
  \caption{Qualitative comparison on representative degraded images. Red boxes on the LQ inputs identify the enlarged regions shown for each method.}
  \label{fig:qualitative_comparison}
\end{figure}

\textbf{Qualitative comparison.}
Figure~\ref{fig:qualitative_comparison} shows that RegionAgent suppresses residual degradations while producing clear local structures. In the landscape, it reduces rain streaks that remain in PromptIR, AdaIR, and RAR. In the building facade, RegionAgent produces straight boundaries and distinct cladding lines, whereas RAR introduces visibly wavy edges.

\subsection{Ablation Study}
\label{sec:ablation_study}

\Needspace{4\baselineskip}
\textbf{Preference optimization.}
We evaluate the effects of two-stage DPO and supervision scope on restoration quality, keeping the shared restorer, expert checkpoints, Localizer, fusion module, and inference budget fixed. A uses SFT alone. B and C add step-level DPO with whole-action and differing-component supervision, respectively. D applies trajectory-level DPO directly after SFT. E and F start from C and add trajectory-level DPO: E supervises the suffix beginning at the differing action, whereas F supervises that action alone and constitutes the full method.

\begin{table}[!htbp]
\centering
\caption{Ablation studies on MiO100. We compare policy training strategies on Group C and expert composition, spatial masks, and Laplacian fusion on Group A.}
\label{tab:ablation}
\begin{minipage}[t]{0.49\linewidth}
\vspace{0pt}
\begingroup
\fontsize{8}{10}\selectfont
\setlength{\tabcolsep}{1.4pt}
\renewcommand{\arraystretch}{1.08}
\begin{tabular*}{\linewidth}{@{\extracolsep{\fill}}crrrrr@{}}
\toprule
\multirow{2}{*}{\textbf{Variant}} & \multicolumn{4}{c}{Restoration quality} & \multicolumn{1}{c}{\multirow{2}{*}{$N_{\rm ref}$}} \\
\cmidrule(lr){2-5}
& \textbf{PSNR}$\uparrow$ & \textbf{SSIM}$\uparrow$ & \textbf{LPIPS}$\downarrow$ & \textbf{MUSIQ}$\uparrow$ & \\
\midrule
A & 20.01 & 0.6264 & 0.3341 & 59.32 & 7.79 \\
B & 20.25 & 0.6389 & 0.3260 & 60.66 & 7.63 \\
C & 20.47 & 0.6469 & 0.3138 & 62.81 & 7.61 \\
D & 20.37 & 0.6437 & 0.3149 & 62.47 & 6.48 \\
E & 20.69 & 0.6544 & 0.3078 & 63.24 & 7.34 \\
\textbf{F} & \textbf{20.90} & \textbf{0.6591} & \textbf{0.2976} & \textbf{64.29} & 7.12 \\
\bottomrule
\end{tabular*}
\endgroup
\end{minipage}\hfill
\begin{minipage}[t]{0.49\linewidth}
\vspace{0pt}
\begingroup
\fontsize{8}{10}\selectfont
\setlength{\tabcolsep}{1.35pt}
\renewcommand{\arraystretch}{1.08}
\begin{tabular*}{\linewidth}{@{\extracolsep{\fill}}lrrrr@{}}
\toprule
\multicolumn{1}{c}{\multirow{2}{*}{\textbf{Variant}}} & \multicolumn{2}{c}{Final image} & \multicolumn{2}{c}{Fusion error} \\
\cmidrule(lr){2-3}\cmidrule(l){4-5}
& \textbf{PSNR}$\uparrow$ & \textbf{LPIPS}$\downarrow$ & $E_{\rm keep}\!\downarrow$ & $E_{\rm bnd}\!\downarrow$ \\
\midrule
Single expert & 21.59 & 0.2537 & 0.61 & 4.12 \\
w/o $M_{\rm geo}$ & 21.91 & 0.2433 & 1.38 & 4.69 \\
w/o $M_{\rm fail}$ & 21.82 & 0.2450 & 2.43 & 4.40 \\
w/o $M_{\rm edge}$ & 22.02 & 0.2432 & 1.02 & 6.28 \\
w/o Laplacian pyramid & 21.83 & 0.2445 & 1.31 & 4.98 \\
\textbf{Full method} & \textbf{22.11} & \textbf{0.2377} & 0.85 & 3.28 \\
\bottomrule
\end{tabular*}
\endgroup
\end{minipage}
\par\vspace{4pt}
\begin{minipage}[t]{0.49\linewidth}
\raggedright\footnotesize\textbf{(a) Preference optimization.} $N_{\rm ref}$ denotes the mean refinement count.
\end{minipage}\hfill
\begin{minipage}[t]{0.49\linewidth}
\raggedright\footnotesize\textbf{(b) Local fusion.} $E_{\rm keep}$ and $E_{\rm bnd}$ are reported in units of $10^{-3}$.
\end{minipage}
\end{table}

F improves PSNR over SFT alone (A) by 0.89~dB and over the single-stage variants C and D by 0.43 and 0.53~dB, respectively, achieving the best restoration quality among the evaluated variants (Table~\ref{tab:ablation}(a)). F also reduces the mean \texttt{REFINE} count from C's 7.61 to 7.12.

With the same action pairs and update budget, C exceeds B by 0.22~dB and reduces LPIPS from 0.3260 to 0.3138, supporting supervision focused on the differing configuration component. Starting from the same step-level checkpoint and using the same trajectory pairs and update budget, F exceeds E by 0.21~dB. This favors assigning trajectory preferences to the sole differing action over the complete suffix.

\textbf{Local fusion.}
In Table~\ref{tab:ablation}(b), the full method improves PSNR by 0.52~dB over single-expert execution and achieves lower LPIPS. To evaluate the masks and multiscale blending, we compare different fusion variants using the same current image, selected region, restored crop, and fusion strength. Removing $M_{\rm fail}$ increases the added error in pixels well restored by Base ($E_{\rm keep}$) from 0.85 to 2.43. Without $M_{\rm edge}$, boundary gradient error ($E_{\rm bnd}$) rises from 3.28 to 6.28; it also increases to 4.69 without $M_{\rm geo}$ and 4.98 without the Laplacian pyramid.
\FloatBarrier

\section{Conclusion}
\label{sec:conclusion}

We present RegionAgent, an adaptive restoration framework that uses a VLM to generate region-specific restoration actions for images with non-uniform coupled degradations. To address the lack of preferences over regional restoration decisions in conventional image-pair datasets, we construct a dataset of action and trajectory preference pairs. Using these pairs, step-level DPO first trains the VLM to select configurations suited to current regional needs. Trajectory-level DPO then uses complete-trajectory feedback to train the VLM to account for the effects of current actions on subsequent restoration. Experiments across multiple datasets demonstrate strong restoration performance under both mixed and single degradations, while ablation studies validate the effectiveness of two-stage preference learning.

\bibliographystyle{iclr2027_conference}
\bibliography{references}

\end{document}